\documentclass[11pt]{article}

\usepackage[preprint]{acl}

\usepackage{times}
\usepackage{latexsym}

\usepackage[T1]{fontenc}

\usepackage[utf8]{inputenc}

\usepackage{microtype}

\usepackage{inconsolata}

\usepackage{graphicx}

\usepackage{amsmath}
\usepackage{booktabs}

\title{The Language of Looking: Motion Tokens Reveal Domain-Dependent Local Structure in Gaze Sequences
}

\author{
  Virmarie Maquiling \and Enkelejda Kasneci \\
  Human-Centered Technologies for Learning \\
  Technical University of Munich \\
  Munich Center for Machine Learning (MCML) \\
  Munich, Germany \\
  \texttt{\{virmarie.maquiling,enkelejda.kasneci\}@tum.de}
}

\hypersetup{
  pdftitle={The Language of Looking: Motion Tokens Reveal Domain-Dependent Local Structure in Gaze Sequences},
  pdfauthor={Virmarie Maquiling and Enkelejda Kasneci},
  pdfsubject={Accepted to Findings of EMNLP 2026}
}

\begin{document}
\maketitle
\begin{abstract}
Human gaze is the primary high-bandwidth interface through which we parse the semantics of our environment. While often modeled as a sequence, the degree to which gaze exhibits transferable local structure across domains remains underexplored. Using motion tokenization as a diagnostic lens across four diverse datasets, we investigate the transferability of local sequence structure. We demonstrate that while gaze can share a reusable motion-token vocabulary, identical motion tokens are organized into different local transition patterns depending on the dataset and task context. Our results show the limitations of naive data-pooling under low-semantic motion-token representations and provide a framework for evaluating behavioral sequences through information-theoretic diagnostics.

\end{abstract}

\section{Introduction}

Human gaze is more than a biological reflex, it is the high-bandwidth serialization of human perception. Just as natural language represents the serialization of thought into discrete symbols, gaze represents the serialization of environmental input into an ordered stream of information. This ``Language of Looking'' has long been recognized as structured and task-driven. More specifically, early work by \citeauthor{yarbus1967eye} reported that viewing patterns shift according to task instructions \cite{yarbus1967eye}, while \citeauthor{noton1971scanpaths} formalized these sequences as ``scanpaths'' composed of stable fixations and rapid saccades \cite{noton1971scanpathsb,noton1971scanpaths}. This sequential view of gaze motivates examining whether the organization of gaze remains consistent when represented and modeled across different domains.

Recent work represents gaze sequentially in multimodal models for scanpath prediction and generation~\cite{mondal2023gazeformer,bolliger2023scandl}, gaze-augmented language understanding~\cite{deng2023pre}, and reading-goal inference \cite{shubi2025decoding}. These applications motivate a basic transfer question: \textbf{when gaze is represented as a token sequence, which short-range regularities are reusable across domains?}

The continuous gaze signal does not provide a unique, predefined token vocabulary. Its units depend on how the signal is represented. We therefore treat fixation-to-fixation motion as the diagnostic unit: each exported sequence begins with an alignment entry, after which each token represents a local gaze displacement. The resulting sequences expose short-range transition patterns between movements. Gaze datasets are collected under heterogeneous regimes--headworn everyday activities, in-the-wild reading, screen-based visual search, and sentence reading. If these regimes share compatible transition structure, pooled training may exploit common dynamics. If their transition patterns are domain-dependent, however, pooling strategy becomes consequential, because unconditional aggregation may obscure dataset-specific regularities even when the token vocabulary is shared.

We investigate this structural compatibility using a motion-based tokenization pipeline. More specifically, using two primary tokenizers---\texttt{AngDelta} (a deterministic geometric grid) and \texttt{VQDelta} (a learned vector-quantized vocabulary)---we unify four datasets spanning headworn (AEA \cite{lv2024aria}, RITW \cite{yang2026reading}) and screen-based (COCO-Search18TP \cite{chen2021coco}, ZuCo 2.0 \cite{hollenstein2020zuco}) regimes. Our analysis moves beyond predictive accuracy to include information-theoretic diagnostics:

\begin{itemize}
    \item \textbf{Next-token transfer and order sensitivity} to measure the reliance on local context.
    \item \textbf{Motif overlap} (Jaccard similarity) of bigrams and trigrams to identify shared local motion patterns.
    \item \textbf{Shared-token transition analysis} to distinguish shared token inventories from shared transition structure.
\end{itemize}

Our findings show that \textit{shared tokens do not imply shared transitions}. Datasets can use overlapping motion-token vocabularies while organizing them into different local patterns. Across the studied datasets, short-range, order-sensitive transition structure transfers unevenly, and apparent compatibility depends on the tokenizer and source--target pair. These results motivate evaluating token overlap and transition structure separately in cross-dataset gaze models.

\section{Related Work}
\subsection{Scanpath Representations}
Eye movements are often represented as ordered sequences of fixations and saccades, commonly referred to as scanpaths~\cite{noton1971scanpathsb, noton1971scanpaths}, and are shaped by factors such as task, stimulus structure, and viewing context~\cite{foulsham2008turning, marius2009gaze, land2006eye}. However, scanpath structure is not independent of representation. Different representations preserve different aspects of gaze. For example, ScanMatch~\cite{cristino2010scanmatch} converts fixations into symbolic sequences for alignment-based comparison, while MultiMatch~\cite{dewhurst2012depends, jarodzka2010vector} avoids a single similarity score by representing scanpaths as vectors and decomposing similarity into shape, saccade direction and length, fixation position, and fixation duration. Comparative studies show that scanpath metrics quantify different aspects of gaze behavior, and therefore need to be chosen according to the structure one wants to preserve~\cite{anderson2015comparison}. The effects of scanpath feature engineering have also been examined in supervised image-classification settings~\cite{byrne2023exploring}. More recent reviews emphasize that scanpath analysis depends heavily on representational choices, such as how continuous gaze coordinates are discretized, how fine the spatial bins are, and whether fixation duration is encoded~\cite{laborde2026vision}. This makes representation choice central when asking whether gaze structure transfers across datasets. A similar issue arises in computational psycholinguistics, where tokenization is treated as an algorithmic intermediate step whose units can misalign with experimentally defined regions of interest~\cite{giulianelli2024proper}. We take the same view of gaze tokenization: motion-based tokens are a measurement lens for exposing one specific aspect of gaze, namely local fixation-to-fixation transition structure.

\subsection{Local motifs/n-grams/Markov structure and transition dynamics}
Our work is also related to approaches that model gaze as local transitions, motifs, or subsequences. Hidden Markov Models and related methods represent gaze as transitions between observed or latent states, often to characterize viewing strategies or infer task structure~\cite{chuk2014understanding, chuk2017having, chuk2020eye, coutrot2018scanpath}. Recurrence-based analyses similarly treat scanpaths as temporally structured behavior, capturing repeated subsequences, revisits, and other order-sensitive regularities~\cite{anderson2013recurrence, vaidyanathan2014recurrence}. Other work explicitly models short subsequences. Methods like SubsMatch represent scanpaths using frequencies of short patterns, making local order explicit~\cite{kubler2017subsmatch}. Related n-gram analyses show that short motifs can be informative, while their frequencies decrease rapidly with length and the most useful n-gram length varies across datasets~\cite{reani2018investigation}.

This literature shows that local transition is a meaningful object of study. However, most work focuses on structure within a dataset or task. The open question now is whether these local patterns transfer across domains. Our work addresses this by asking whether motion-based transition patterns are shared, partially shared, or regime-specific across datasets.

Information-theoretic work has also examined whether gaze follows entropy-rate constancy, a principle originally proposed for text by \citet{genzel2002entropy, genzel2003variation}. \citet{wang2023does} and \citet{wang2024much} investigate temporal information-density patterns in listener gaze during face-to-face interaction. Our analysis asks a different question: instead of testing whether information density remains constant over an interaction, we use target regret, order controls, motif overlap, and conditional transition entropy to test whether fixation-motion transition structure transfers across datasets.

\subsection{Sequence modeling, representation, and domain shift}
Deep learning models increasingly represent gaze as a sequence to be predicted, generated, or used as supervision. Models such as Eyettention~\cite{deng2023eyettention}, ScanEZ~\cite{sood2025scanez}, and ScanDL~\cite{bolliger2023scandl} generate or predict scanpaths using sequence models and evaluate both within and across datasets. Other approaches model gaze for visual search, object-level attention, or multimodal reasoning, including transformer-based models and gaze-supervised vision-language models~\cite{mondal2023gazeformer, fang2024oat, li2026thinking}. Related work on gaze tokenization shows that discretization choices affect downstream sequence modeling performance~\cite{rolff2025tokenization}. \citeauthor{maquiling2026motionbased} studied motion-based gaze tokenization primarily as a representation for cross-dataset transfer in egocentric eye-movement data~\cite{maquiling2026motionbased}. In contrast, this work uses tokenized gaze sequences as a diagnostic tool to test whether local transition structure is reusable across broader dataset regimes, including both headworn and screen-based settings.

At the same time, domain shift is a well-known issue in gaze modeling. Cross-dataset degradation has been studied extensively in gaze estimation and is increasingly considered in scanpath prediction and reading models~\cite{zhang2015appearance, kellnhofer2019gaze360, sood2025scanez, deng2023eyettention}. However, most work evaluates transfer in terms of predictive performance--such as likelihood, scanpath similarity, or task accuracy.

What is less studied is whether the underlying sequence structure itself transfers across datasets and regimes. Prior work already treats gaze as a discretized sequence with informative local transitions. We instead ask whether the local structure exposed by shared motion tokens is reusable across datasets. A model may transfer because token frequencies overlap, even if local transition patterns differ. We therefore combine transfer metrics with order sensitivity, motif overlap, and transition analysis to distinguish token overlap from structural overlap, testing whether gaze, represented as fixation-level motion tokens, shares common local structure or forms domain-dependent transition patterns.

\section{Methodology}

\subsection{Datasets and Regimes}

We study whether local sequential structure in gaze transfers across recording and task regimes. For this, we use two headworn eye-tracking datasets, ARIA Everyday Activities~\cite{lv2024aria} and Reading Recognition in the Wild--Columbus Set ~\cite{yang2026reading}, and two screen-based datasets, COCO-Search18TP~\cite{chen2021coco} and ZuCo 2.0~\cite{hollenstein2020zuco}. These datasets--shortened in this paper to AEA, RITW, COCO, and ZuCo, respectively--span two broad recording regimes: headworn gaze, recorded with wearable eye tracking devices, and screen-based gaze, recorded with remote eye trackers during screen viewing. They also differ in task structure: general activities, reading recognition in the wild, visual search, and sentence reading, respectively.

We group directed transfers into four regime families: within-headworn (AEA$\rightarrow$RITW and RITW$\rightarrow$AEA), within-screen (COCO$\rightarrow$ZuCo and ZuCo$\rightarrow$COCO), headworn$\rightarrow$screen (AEA/RITW$\rightarrow$COCO/ZuCo), and screen$\rightarrow$headworn (COCO/ZuCo$\rightarrow$AEA/RITW). We use ``regime'' descriptively because task, stimulus, recording setup, and preprocessing co-vary in this benchmark. This helps us examine whether transfer quality also reflects broader recording and task conditions.

\subsection{Event Representation}

To compare datasets consistently, we represent all gaze data as fixation sequences. AEA and RITW, which are recorded as continuous gaze streams, are segmented into events using a simplified adaptive velocity-threshold identification (I-VT) procedure conceptually based on~\citeauthor{nystrom2010adaptive}~(\citeyear{nystrom2010adaptive}), with a median absolute deviation (MAD)-based per-sequence velocity threshold. Only fixation events are retained. COCO and ZuCo are already provided as fixation sequences and are used directly.

\subsection{Motion Tokenizers}
We focus the main analysis on two motion-based tokenizers.

\paragraph{AngDelta.}
AngDelta is a deterministic fixation-to-fixation motion tokenizer. For each adjacent fixation pair, it computes the pseudo-angular change $(\Delta \mathrm{yaw}, \Delta \mathrm{pitch})$ and quantizes that 2D motion on a fixed $[-120^\circ, 120^\circ]\times[-120^\circ, 120^\circ]$ range using a $24\times18$ uniform grid. The mapping is shared across datasets, which makes token identities directly comparable across regimes.
\paragraph{VQDelta.}
VQDelta is a vector-quantized motion tokenizer. It computes the same fixation-to-fixation angular deltas, but assigns each delta to the nearest centroid in a learned codebook. For the shared-vocabulary reference, we fit one $K=128$ codebook to normalized motion deltas pooled from the training splits of AEA, RITW, COCO, and ZuCo. The codebook is then frozen and reused for every directed transfer. Because the target training split contributes to this pooled codebook, this condition tests transfer under a shared learned vocabulary as opposed to strict target-unseen tokenization. We therefore also evaluate source-only and leave-target-out $K=128$ codebooks in Appendix~\ref{app:additional_analyses}.

\begin{figure}[t]
    \centering
    \includegraphics[width=.8\columnwidth]
    {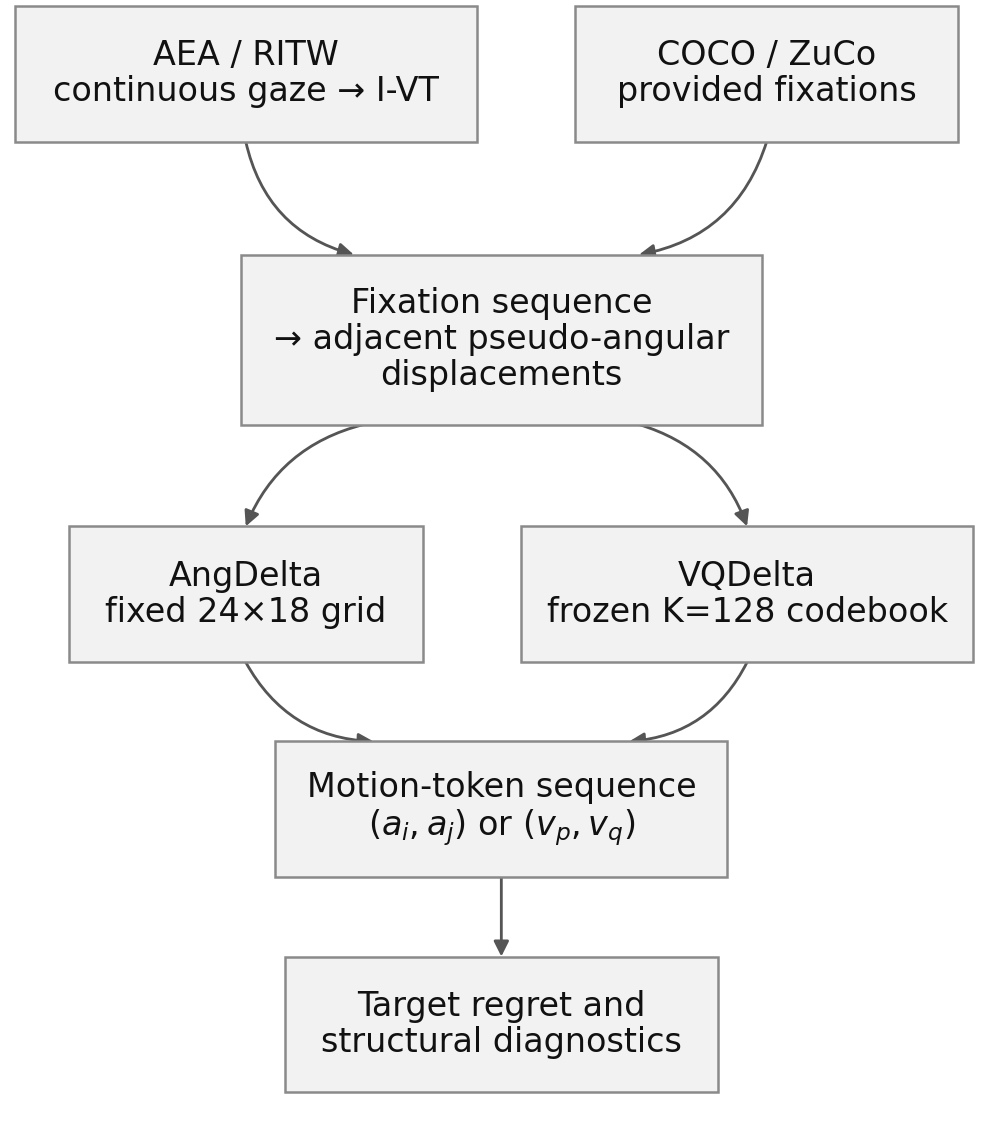}
    \caption{Motion-tokenization pipeline. Continuous AEA and RITW recordings are segmented into fixations using adaptive I-VT, whereas COCO and ZuCo provide fixation sequences directly. Adjacent fixations are converted into pseudo-angular displacements and quantized using either the fixed AngDelta grid or a frozen VQDelta codebook. For three fixations $(f_1,f_2,f_3)$, the resulting displacements become token sequence $(a_i,a_j)$ under AngDelta or $(v_p,v_q)$ under VQDelta. Screen-based pseudo-angles represent normalized screen displacement, not physical eye rotation. The token sequences are used for target-regret evaluation, order controls, motif overlap, and transition diagnostics. Initial alignment entries are omitted for clarity.}
    \label{fig:tokenization_pipeline}
\end{figure}

Figure~\ref{fig:tokenization_pipeline} summarizes the dataset-specific preprocessing paths, the two tokenization procedures, and the resulting motion-token sequences.

AngDelta serves as the main interpretable structural probe, while VQDelta tests whether the regime-level pattern persists under a learned motion vocabulary. Both representations rely on angular changes in gaze. While these are directly available in headworn datasets, screen-based datasets provide gaze in pixel coordinates. To ensure compatibility, we convert screen-based fixations into normalized pseudo-angles by rescaling coordinates and mapping them to unit rays. For screen datasets, pseudo-angles are not physical eye rotations but normalized relative screen-motion coordinates. Cross-regime comparisons therefore test shared local displacement structure rather than identical underlying gaze rotations.

\subsection{Transfer Evaluation}

For each tokenizer, we export all four datasets using the same configuration and keep the tokenizer fixed across all transfer directions. We train an add-$k$ smoothed order-3 n-gram model on source-training sequences and evaluate it on held-out target sequences. Our primary transfer metric is target regret:
\begin{equation}
\mathcal{R}_{s\rightarrow t}
=
H_t(p_s)-H_t(p_t),
\end{equation}
where $H_t(p_s)$ is the source-trained model's cross-entropy on the target test sequences and $H_t(p_t)$ is the cross-entropy of a target-trained reference evaluated on the same sequences. Both are measured in bits per token. Positive regret indicates that the source-trained model predicts the target data less well than the target-trained reference. The reference is a finite-sample comparison model, not an oracle or mathematical upper bound. We also report in-domain and out-of-domain perplexity and the source-relative degradation $(\mathrm{OOD}-\mathrm{IND})/\mathrm{IND}$ as secondary descriptive quantities in Appendix~\ref{app:uncertainty}. Full implementation details are provided in Appendix~\ref{app:impl_details}.

To test whether the model is actually learning sequence structure as opposed to just token marginals, we use an order-manipulation control on the target side. We first evaluate the source-trained model on the completed export target sequences, then reevaluate it after shuffling all entries within each sequence. We define order sensitivity as the relative increase in OOD perplexity under this target-side shuffle:
\begin{equation}
\mathrm{Order\ Sensitivity} =
\frac{\mathrm{PPL}_{\text{shuf}} - \mathrm{PPL}_{\text{real}}}{\mathrm{PPL}_{\text{real}}}
\end{equation}
where $\mathrm{PPL}_{\text{real}}$ is the perplexity on original target sequences and $\mathrm{PPL}_{\text{shuf}}$ is the perplexity after shuffling tokens within each sequence.

Order sensitivity therefore measures how much destroying order at evaluation harms prediction under the same source-trained model. Because shuffling preserves the target token inventory while disrupting local order, this control helps distinguish transfer based on token marginals from transfer based on sequential structure.

\paragraph{N-gram and shuffle controls.}
We compare n-gram orders 1--4 under the same token streams and target-trained-reference protocol. Order 1 provides a unigram baseline that contains no sequential context. In addition to the relative-perplexity order-sensitivity measure above, we report the absolute shuffled-minus-real increase in target cross-entropy, in bits per token, after independently shuffling tokens within each target sequence. These are related but distinct diagnostics.

\paragraph{Leave-target-out pooling.}
For each target dataset, we combine the remaining three datasets using three strategies: naive count pooling, dataset-balanced count pooling, and an equal-weight probability mixture of three separately trained source models. The probability mixture has no target-conditioned gate. Each pooled model is compared with the individual source models and with the target-trained reference on the same target test sequences.

\paragraph{Stability analyses.}
At split seed 0, we assess sequence-level uncertainty using 1000 bootstrap resamples of complete target sequences. We also repeat scoring after progressively excluding short sequences and under natural-context-only evaluation, which scores only positions with enough observed preceding tokens for the full n-gram context. These analyses test sensitivity to small target partitions, short sequences, and sequence-start padding.

\subsection{Structural Diagnostics}

Perplexity alone does not reveal whether two datasets share the same local sequence structure. A model may transfer because frequent tokens overlap, even if those tokens are arranged into different local patterns. We therefore complement transfer perplexity with token- and motif-level diagnostics computed from the exported token streams: top-token overlap, top-200 bigram and trigram Jaccard overlap, and pair-specific transition matrices over shared high-frequency tokens. These diagnostics separate whether datasets share the same motion-token vocabulary from whether they organize those tokens in similar transitions.

\subsection{Structural Predictability via Transition Entropy}
To quantify within-domain local predictability, we compute the first-order conditional entropy $H(X_{\text{next}} | X_{\text{current}})$. This metric measures the uncertainty of the next motion token given the current token:

\begin{equation}
H(X_{\text{next}} | X_{\text{current}}) = - \sum_{x,y \in V} P(x,y) \log_2 P(y|x)
\end{equation}

where $V$ is the observed motion-token vocabulary, $P(x,y)$ is the empirical bigram probability, and $P(y\mid x)$ is the row-normalized transition probability. Because datasets can differ in the number of motion tokens they use, we also report a version normalized by the observed vocabulary size:
\begin{equation}
\bar{H}(X_{\text{next}} \mid X_{\text{current}}) =
\frac{H(X_{\text{next}} \mid X_{\text{current}})}
{\log_2 |V_{\mathrm{used}}|}.
\end{equation}
Lower conditional entropy indicates more concentrated next-token distributions, whereas higher conditional entropy indicates more diffuse local transitions. Together with pairwise transfer and motif overlap, this metric characterizes variation in local transition structure across datasets.

\subsection{Controls: Length and Event-interface Effects}

We include two controls to test whether the observed grouping patterns can be explained by artifacts of sequence construction. First, the matched-length control repeats the main fixation-level transfer evaluation after truncating all sequences to a common maximum length of 12 tokens. This tests whether regime differences are driven by unequal sequence lengths under the same fixation-level interface. Second, the mixed-format event-interface control compares the unified fixation-level benchmark against a setting where AEA and RITW retain full event streams while COCO and ZuCo remain fixation-native. This tests how much cross-regime degradation is inflated by incompatible event interfaces. Full details and results are provided in Appendix~\ref{app:controls}.

\subsection{Availability}

Experiments were run on a Windows 11 workstation with an NVIDIA RTX 4090 GPU. All tokenization, training, and evaluation code are available in the following repository: \url{https://github.com/vbmaq/language-of-looking/}. The release includes the exact target-regret, pooling, VQDelta-codebook, and stability configurations; split-seed and execution manifests; and provenance records for the reported summary tables.

\section{Results}

\subsection{Local Structure Transfers Unevenly Across Datasets}

\begin{figure*}[ht]
    \centering
    \includegraphics[width=0.85\linewidth]{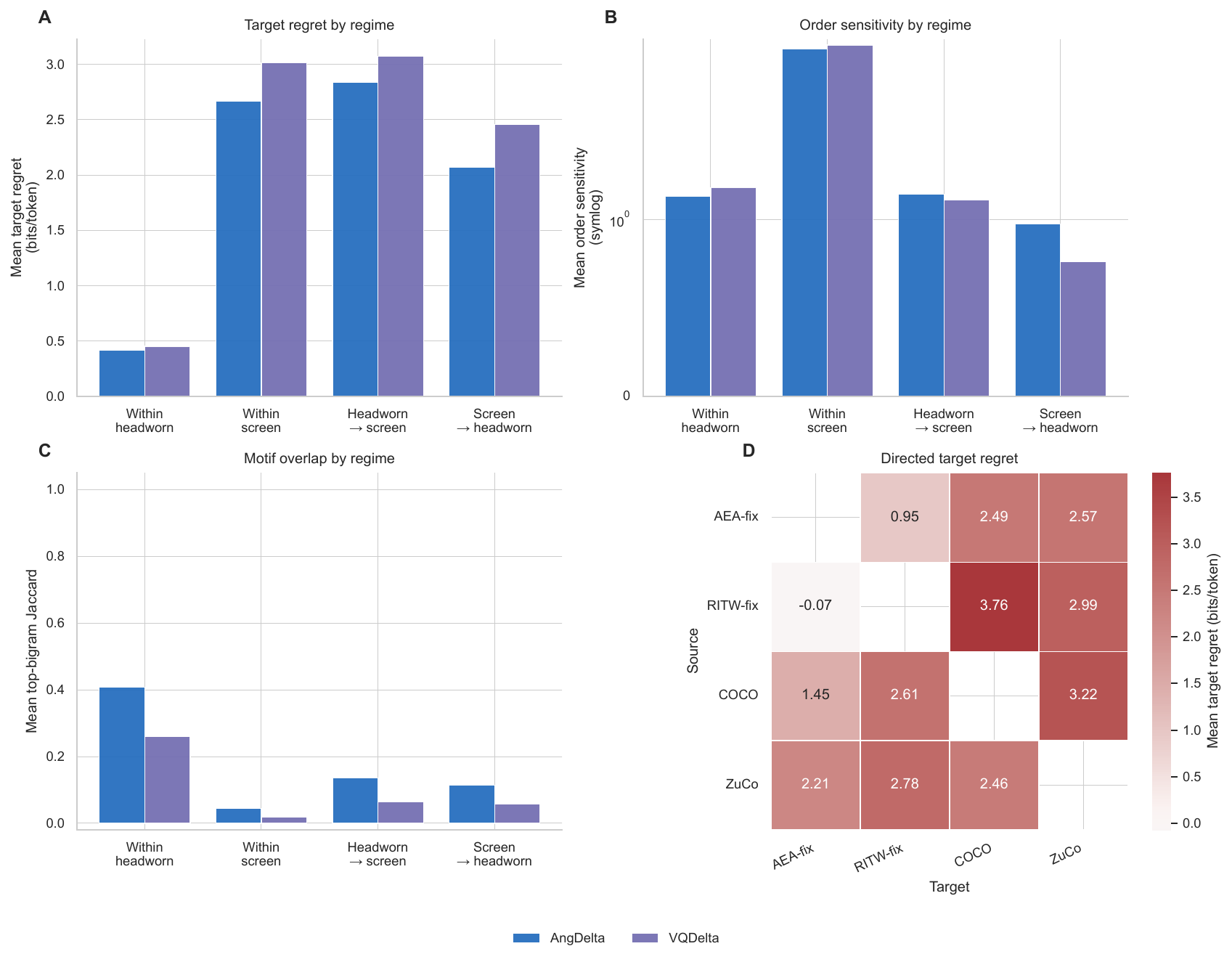}
    \caption{Cross-dataset transfer and structural diagnostics under the unified fixation-level benchmark. Panel A reports mean order-3 target regret across five split seeds. Positive values indicate worse prediction than a target-trained reference evaluated on the same target sequences. Panels B and C report the fixed-seed order-sensitivity and bigram-overlap diagnostics. Panel D shows directed target regret averaged across both tokenizers and five split seeds. The ``-fix'' suffix indicates that non-fixation events were removed from AEA and RITW to match the fixation-native screen datasets.}
    \label{fig:regime_transfer}
\end{figure*}

Figure~\ref{fig:regime_transfer}A summarizes order-3 target regret across five split seeds. Averaged over 60 direction--seed evaluations per tokenizer, mean regret was $2.150$ bits/token for AngDelta and $2.420$ for VQDelta. Regret was positive in all 24 evaluations for split seed 0 and in 57/60 AngDelta and 58/60 VQDelta evaluations across the five seeds. Thus, source-trained models generally predicted target sequences less well than target-trained references evaluated on the same target data.

The descriptive dataset groupings did not show a general within-group advantage. Mean AngDelta/VQDelta regret was $0.421/0.453$ bits/token within-headworn, $2.667/3.015$ within-screen, and $2.453/2.763$ across the two recording groups. The relatively low within-headworn average was partly driven by the weak RITW$\rightarrow$AEA direction. Sequence-level bootstrapping gave regret of $0.038$ bits/token for AngDelta with 95\% CI $[-0.307,0.465]$ and $0.133$ for VQDelta with 95\% CI $[-0.118,0.401]$. Only six target-AEA test sequences remain after requiring length 12, and natural-context-only RITW$\rightarrow$AEA regret becomes slightly negative ($-0.087$ for AngDelta and $-0.017$ for VQDelta). COCO has median sequence length 3, and only 37.3\% of its order-3 predictions have a complete two-token context. At minimum length 12, direction-ranking correlations with the length-2 evaluation fall to $0.552$ for AngDelta and $0.559$ for VQDelta. We therefore treat RITW$\rightarrow$AEA as uncertain, restrict COCO conclusions to very short-range structure, and avoid interpreting the exact direction ranking as stable.

\subsection{Order Sensitivity Shows That Transfer Depends on Sequence Structure}
Figure~\ref{fig:regime_transfer}B reports the fixed-seed relative increase in target perplexity after shuffling tokens within each sequence. For AngDelta, this value is $1.23$ within-headworn, $4.40$ within-screen, $1.24$ for headworn$\rightarrow$screen, and $0.98$ for screen$\rightarrow$headworn. VQDelta gives $1.32$, $4.56$, $1.18$, and $0.76$ across the same descriptive groupings. N-gram-order and shuffle controls support order sensitivity beyond unigram frequencies. Relative to order 1, the order-3 configuration reduces aggregate target cross-entropy by $0.392$ bits/token for AngDelta and $0.917$ for VQDelta. Shuffling increases target cross-entropy in all 24 directions for split seed 0, by an average of $1.188$ bits/token for AngDelta and $1.113$ for VQDelta. Order 2 performs best overall for AngDelta, whereas orders 3 and 4 are nearly tied for VQDelta. These results support short-range order sensitivity, but do not imply linguistic syntax or that order 3 is universally optimal.

\paragraph{Pooling and codebook controls.}
Across the eight target--tokenizer conditions, the equal-weight source-model mixture was the best multi-source strategy in six conditions, dataset-balanced pooling in two, and naive pooling in none. The best multi-source strategy improved over the average of the three single-source models, although it did not always outperform the strongest individual source. It remained worse than the target-trained reference in seven of eight conditions. The exception was AEA--AngDelta with balanced pooling, with regret $-0.113$ bits/token. For VQDelta at $K=128$ and split seed 0, mean regret was $2.344$ bits/token with the pooled shared-vocabulary codebook, $2.232$ with source-only codebooks, and $2.283$ with leave-target-out codebooks. The comparable means indicate that VQDelta regret under the pooled codebook is not explained solely by target exposure during codebook fitting, although refitting the codebook changes the discretization as well as the transition model. Appendix~\ref{app:additional_analyses} reports the full control summaries.

\subsection{Local Motif Overlap Reveals Which Regimes Share Sequence Structure}

\begin{figure*}[ht]
    \centering
    \includegraphics[width=.8\linewidth]{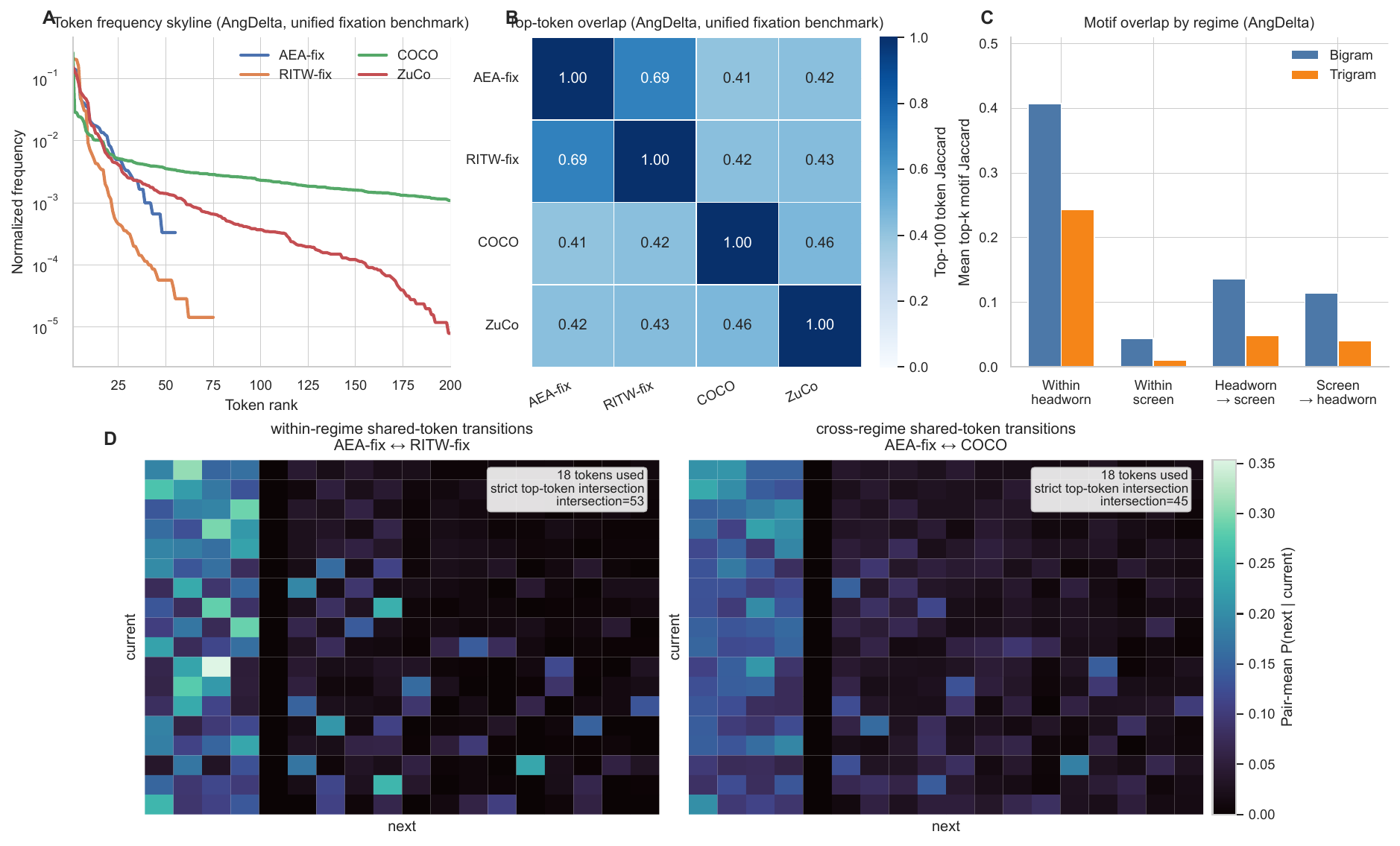}
    \caption{Interpretable local structure under AngDelta. Panel D averages row-normalized transitions for each dataset pair after projecting onto pair-specific shared top tokens. Shared tokens can therefore induce different local transition systems across regimes. For readability, an 18-token subset of the shared set is shown.}
    \label{fig:angdelta}
\end{figure*}

Figure~\ref{fig:angdelta} shows how tokens and local token structure are shared across datasets. Panel~C summarizes regime-averaged motif overlap. Within-headworn overlap is the strongest, with mean bigram and trigram Jaccard values of $0.41$ and $0.24$. Within-screen COCO$\leftrightarrow$ZuCo overlap is much lower ($0.04$ and $0.01$), despite both datasets being fixation-native screen recordings. This suggests that screen-based search and sentence reading are operationally compatible as fixation sequences, but they do not share the same local motion motifs to the same extent as the two headworn datasets. We use top-200 motifs in the main analysis. Appendix~\ref{app:topk_sensitivity} shows that the broad regime ordering is stable across top-$k$ thresholds from 25 to 500.

Looking beyond the regime averages, directed motif values show that cross-regime overlap is pair-specific. AEA$\rightarrow$COCO reaches only $0.06$ bigram Jaccard and $0.02$ trigram Jaccard, whereas AEA$\rightarrow$ZuCo reaches $0.19$ and $0.07$. The RITW pattern is stronger for ZuCo than for COCO: RITW$\rightarrow$ZuCo reaches $0.24$ bigram Jaccard and $0.10$ trigram Jaccard, while RITW$\rightarrow$COCO remains at $0.06$ and $0.02$. %
Descriptively, these values show that cross-group overlap is not uniformly absent and varies across the specific dataset pair. Because motif sets were not bootstrapped by resampling complete sequences, we treat these overlaps as supporting diagnostics.%

Figure~\ref{fig:angdelta}D makes this clearer. In the within-headworn AEA/RITW comparison, transitions are concentrated: given a token, only a few next tokens dominate. In the cross-regime AEA$\leftrightarrow$COCO case, transitions are more diffuse and scattered. %
The comparison shows that overlapping token inventories can accompany different empirical transition matrices. It does not identify which task, stimulus, device, or preprocessing factor causes those differences.

\subsection{Transition Entropy Quantifies Local Predictability}
\label{res:transition_entropy}

To complement motif-overlap and transition-matrix analyses, we compute raw and normalized first-order transition entropy per dataset and tokenizer. Raw conditional entropy measures next-token uncertainty in bits, while the normalized value normalizes by observed vocabulary size. Lower values indicate more constrained local transitions.

\begin{table}[ht]
\centering
\small
\begin{tabular}{llrr}
\toprule
Tokenizer & Dataset & $H(X_t \mid X_{t-1})$ & Norm. $H$ \\
\midrule
AngDelta & AEA-fix  & 3.37 & 0.58 \\
AngDelta & RITW-fix & 2.79 & 0.45 \\
AngDelta & COCO     & 5.97 & 0.77 \\
AngDelta & ZuCo     & 3.75 & 0.48 \\
\midrule
VQDelta & AEA-fix  & 4.15 & 0.69 \\
VQDelta & RITW-fix & 4.05 & 0.65 \\
VQDelta & COCO     & 5.65 & 0.81 \\
VQDelta & ZuCo     & 4.54 & 0.65 \\
\bottomrule
\end{tabular}
\caption{Local transition predictability across datasets and tokenizers. $H(X_t \mid X_{t-1})$ is the raw first-order conditional entropy in bits. Norm.~$H$ is $H(X_t \mid X_{t-1}) / \log_2 |V_{\mathrm{used}}|$, which partially accounts for differences in observed vocabulary size. Lower values indicate more constrained next-token dynamics.}
\label{tab:transition_entropy}
\end{table}

Table~\ref{tab:transition_entropy} shows that COCO has the most diffuse local transition system across both tokenizers, with the highest raw and normalized conditional entropy. This pattern coincides descriptively with weak within-screen transfer, but does not establish that entropy differences cause the transfer gap. RITW has the lowest raw conditional entropy for both tokenizers, while RITW and ZuCo have the lowest normalized values. This pattern supports the transition-matrix analysis: datasets may share motion-token vocabularies while differing in how predictable their local transitions are.

\subsection{Controls: Length and Event-interface Effects}

The matched-length and mixed-format controls show that source-relative degradation is sensitive to sequence construction. Truncating sequences changes degradation magnitudes, while comparing full headworn event streams against fixation-native screen datasets substantially inflates cross-group degradation. Because degradation depends on the intrinsic predictability of the source and target, these analyses serve as secondary diagnostics of interface and sequence-length effects, not as evidence for a general ordering between regimes. Target-regret stability under minimum-length and natural-context evaluation appears in Appendix~\ref{app:additional_analyses}.

\section{Discussion}
\subsection{Main findings}
Motion tokens reveal reusable local gaze structure, but this structure is not shared equally across domains.

\paragraph{Transfer gaps remain after accounting for target predictability.}
Target regret compares source-trained and target-trained models on the same target sequences, avoiding the intrinsic-predictability confound in source-relative degradation. Both tokenizers show positive regret on average, but transfer reliability varies by source--target pair. In particular, RITW$\rightarrow$AEA remains uncertain under sequence-level bootstrapping, so the results do not support robust transfer in every direction.

\paragraph{Local structure transfers unevenly.}
Transfer varies across source--target pairs and does not follow a simple headworn-versus-screen hierarchy. Because task, stimulus, device, preprocessing, representation, sequence length, and dataset size co-vary, these broad groupings should be treated as descriptive.

\paragraph{Representation changes transfer magnitude.}
Tokenizer choice changes the size of the target-side transfer gap. Mean regret was $2.150$ bits/token for AngDelta and $2.420$ for VQDelta, although the difference varied across source--target pairs. The directional reversal observed under source-relative degradation does not persist under target regret, showing that source-relative normalization can change the apparent transfer asymmetry.

\paragraph{Order matters.}
Order sensitivity shows that the models are not only exploiting token frequencies. Shuffling the target sequences changes prediction performance, so the model is using information from local token order. %

\paragraph{Shared vocabulary does not imply shared transition structure.}

Figure~\ref{fig:angdelta} shows that top-token overlap exists across datasets, including cross-dataset pairs, but motif, transition-matrix, and entropy analyses show that the same tokens are often used differently. Token overlap therefore does not imply shared transition structure. COCO has the highest next-token uncertainty across both tokenizers, while RITW and ZuCo are more locally constrained after accounting for observed vocabulary size. AEA remains relatively constrained, but its rank depends on the tokenizer and normalization.

\paragraph{Cross-dataset structure is pair-specific.}
Motif overlap varies across source--target pairs. AEA$\rightarrow$ZuCo and RITW$\rightarrow$ZuCo show higher overlap than the corresponding comparisons with COCO, indicating that shared local patterns are partial rather than uniform. Because dataset characteristics co-vary, these differences should not be attributed to a single task or recording factor.

\subsection{Practical Implications}

\paragraph{Pooling implications.}
Our direct leave-target-out controls show that no pooling strategy is uniformly best: the equal-weight source-model mixture is best in six of eight target--tokenizer conditions, balanced pooling in two, and naive pooling in none. The best multi-source strategy improves over the average individual source but remains worse than the target-trained reference in seven of eight conditions. These results show that combination strategy matters. They do not imply that unified gaze modeling is infeasible. Neural, adapter-based, target-conditioned, and mixture-of-experts models may represent multiple gaze distributions and remain outside the scope of this count-based diagnostic study.

\paragraph{Tokenizer choice is part of the scientific claim.}
AngDelta is useful because its fixed bins make motif usage and transition matrices interpretable. VQDelta tests whether the same regime-level patterns persist under a learned motion vocabulary, but it can also reshape the apparent asymmetry. This shows that tokenizer choice defines which transfer structure becomes visible. It is not just an implementation detail.

\paragraph{Benchmarks need structure diagnostics.}

We need a better protocol for evaluating gaze representations. Perplexity alone does not reveal whether transfer comes from token overlap or shared sequence structure, and source-relative degradation can confound transfer with differences in intrinsic target predictability. Useful benchmarks should therefore combine target regret with order controls, motif overlap, transition matrices, and controls for sequence length and event format. Source-relative degradation may remain useful as a secondary descriptive quantity. Predictive transfer measures whether a representation generalizes, while structural diagnostics help characterize why.

Overall, motion-token overlap is common across datasets, but the associated local transition structure transfers unevenly and depends on the tokenizer and source--target pair.

\section{Conclusion}
We used motion tokenization to test whether short-range gaze-sequence structure transfers across datasets. Using target-trained references evaluated on the same target sequences, mean order-3 regret was $2.150$ bits/token for AngDelta and $2.420$ for VQDelta. Unigram, n-gram-order, and shuffle controls show that prediction depends on short-range token order, while direct pooling and target-unseen VQDelta controls show that transfer also depends on combination strategy and representation. Transfer gaps remain uneven across source--target pairs: RITW$\rightarrow$AEA is uncertain under sequence-level bootstrapping, AEA-target estimates are sample-limited, and COCO supports only very short-range conclusions. Shared tokens therefore do not imply uniformly transferable local transition structure, but the results do not rule out more flexible unified models.

\section{Limitations}
We evaluate four datasets spanning headworn and screen-based settings, but these do not cover the full range of gaze behavior, recording setups, tasks, or stimuli. %
Because each dataset bundles task, stimulus domain, recording setup, preprocessing, representation, sequence length, and corpus size, the headworn and screen-based groupings are descriptive summaries and do not isolate causal effects of recording setup or task. %
AEA-target estimates are particularly sample-limited: only six target-test sequences remain after requiring length 12. COCO has median sequence length 3, and only 37.3\% of order-3 predictions have a complete two-token context, so its results concern very short-range structure. Minimum-length filtering also changes the exact ranking of transfer directions, and motif overlap was not bootstrapped by resampling complete sequences. We therefore do not claim a stable pairwise hierarchy. %
Future work should test whether the same patterns hold in additional domains and should study which modeling strategies can support better generalization across them. A second limitation is that the screen-based datasets are converted into pseudo-angles. These make screen and headworn datasets comparable under the same motion-token interface, but they should not be interpreted as true physical gaze directions. Moreover, our conclusions are tied to motion tokenization. Motion tokens expose fixation-to-fixation movement structure, but do not model the full oculomotor event stream, including saccade dynamics or fixation durations. Gaze is an information-rich signal, and any tokenization inevitably preserves some structure while discarding others. Other representations may reveal different forms of regularity. Several representation sensitivities remain untested. We did not vary the AngDelta grid resolution or angular range, and the target-unseen VQDelta controls use only $K=128$, codebook seed 0, and split seed 0. VQDelta codebook-size sensitivity and multiseed codebook refitting therefore remain future work. The exported sequences also include an initial alignment entry that is permuted with displacement tokens in the shuffle control. The resulting estimate therefore measures order dependence in the complete exported representation instead of displacement order alone. The present count-based experiments diagnose local sequence structure. They do not demonstrate neural-model performance, downstream gaze-aware utility, or the limits of target-conditioned, adapter-based, or mixture-of-experts models. Finally, this paper focuses on sequence structure while largely abstracting away from scene content. Gaze is not only sequential, it is also strongly shaped by visual semantics, task goals, and stimulus structure. Future work should study how local transition structure interacts with scene information. A natural next step is to integrate motion tokens with visual or semantic scene representations, allowing models to learn both gaze dynamics and the scene content that shapes them.

\section{Ethical Considerations}
Gaze data can contain sensitive personal information, including signals related to attention, behavior, fatigue, and cognitive or health-related traits~\cite{kroger2020does, abdrabou2025gaze}. In educational, workplace, AR/VR, and advertising settings, even coarse gaze tokens could support cognitive-state inference, attention monitoring, profiling, or behavioral re-identification. Discretization may remove some detail, but tokenization should not be treated as anonymization. This work is intended as a structural research diagnostic and is not developed or validated for individual-level inference, employee or student monitoring, surveillance, advertising profiles, or access-control decisions. Any deployment-oriented study should use explicit consent, data minimization, restricted access, retention limits, and evaluation of re-identification and sensitive-inference risks.

We did not collect new human-subject data. All experiments use previously released eye-tracking datasets under their original access conditions, and we rely on the consent procedures, ethics approvals, and data-use terms reported by the original dataset providers. We do not use direct identifiers, infer demographic or protected attributes, or redistribute raw gaze traces. Any released code, models, or derived artifacts preserve the licenses, access restrictions, and redistribution terms of the original datasets.

\bibliography{custom}

@inproceedings{yang2026reading,
 author = {Yang, Charig and Alam, Samiul and Siam, Shakhrul Iman and Proulx, Michael and Mathias, Lambert and Somasundaram, Kiran and Pesqueira, Luis and Fort, James and Sheriffdeen, Sheroze and Parkhi, Omkar and Ren, Yuheng and Zhang, Mi and Chai, Yuning and Newcombe, Richard and Kim, Hyo Jin},
 booktitle = {Advances in Neural Information Processing Systems},
 doi = {10.52202/085713-4728},
 editor = {D. Belgrave and C. Zhang and H. Lin and R. Pascanu and P. Koniusz and M. Ghassemi and N. Chen},
 pages = {141466--141485},
 publisher = {Curran Associates, Inc.},
 title = {Reading Recognition in the Wild},
 url = {https://proceedings.neurips.cc/paper_files/paper/2025/file/cf21f05f83772beeecb4ade4e524076b-Paper-Conference.pdf},
 volume = {38, Main Conference},
 year = {2025}
}

@article{lv2024aria,
  title={Aria everyday activities dataset},
  author={Lv, Zhaoyang and Charron, Nicholas and Moulon, Pierre and Gamino, Alexander and Peng, Cheng and Sweeney, Chris and Miller, Edward and Tang, Huixuan and Meissner, Jeff and Dong, Jing and Somasundaram, Kiran and Pesqueira, Luis and Schwesinger, Mark and Parkhi, Omkar and Gu, Qiao and De Nardi, Renzo and Cheng, Shangyi and Saarinen, Steve and Baiyya, Vijay and Zou, Yuyang and Newcombe, Richard and Engel, Jakob Julian and Pan, Xiaqing and Ren, Carl},
  journal={arXiv preprint arXiv:2402.13349},
  year={2024}
}

@article{cristino2010scanmatch,
  title={ScanMatch: A novel method for comparing fixation sequences},
  author={Cristino, Filipe and Math{\^o}t, Sebastiaan and Theeuwes, Jan and Gilchrist, Iain D},
  journal={Behavior research methods},
  volume={42},
  number={3},
  pages={692--700},
  year={2010},
  publisher={Springer}
}

@article{coutrot2018scanpath,
  title={Scanpath modeling and classification with hidden Markov models},
  author={Coutrot, Antoine and Hsiao, Janet H and Chan, Antoni B},
  journal={Behavior research methods},
  volume={50},
  number={1},
  pages={362--379},
  year={2018},
  publisher={Springer}
}

@article{dewhurst2012depends,
  title={It depends on how you look at it: Scanpath comparison in multiple dimensions with MultiMatch, a vector-based approach},
  author={Dewhurst, Richard and Nystr{\"o}m, Marcus and Jarodzka, Halszka and Foulsham, Tom and Johansson, Roger and Holmqvist, Kenneth},
  journal={Behavior research methods},
  volume={44},
  number={4},
  pages={1079--1100},
  year={2012},
  publisher={Springer}
}

@article{rolff2025tokenization,
  title={Tokenization of Gaze Data},
  author={Rolff, Tim and Karimian, Jurik and Hypki, Niklas and Schmidt, Susanne and Lappe, Markus and Steinicke, Frank},
  journal={arXiv preprint arXiv:2503.22145},
  year={2025}
}

@article{li2026thinking,
  title={Thinking with Gaze: Sequential Eye-Tracking as Visual Reasoning Supervision for Medical VLMs},
  author={Li, Yiwei and Wu, Zihao and Lv, Yanjun and Jiang, Hanqi and You, Weihang and Liu, Zhengliang and Zhu, Dajiang and Li, Xiang and Li, Quanzheng and Liu, Tianming and Zhao, Lin},
  journal={arXiv preprint arXiv:2603.06697},
  year={2026}
}

@article{noton1971scanpaths,
  title={Scanpaths in saccadic eye movements while viewing and recognizing patterns},
  author={Noton, David and Stark, Lawrence},
  journal={Vision research},
  volume={11},
  number={9},
  pages={929--IN8},
  year={1971},
  publisher={Elsevier}
}

@article{noton1971scanpathsb,
  title={Scanpaths in eye movements during pattern perception},
  author={Noton, David and Stark, Lawrence},
  journal={Science},
  volume={171},
  number={3968},
  pages={308--311},
  year={1971},
  publisher={American Association for the Advancement of Science}
}

@article{byrne2023exploring,
  title={Exploring the effects of scanpath feature engineering for supervised image classification models},
  author={Byrne, Sean Anthony and Maquiling, Virmarie and Reynolds, Adam Peter Frederick and Polonio, Luca and Castner, Nora and Kasneci, Enkelejda},
  journal={Proceedings of the ACM on Human-Computer Interaction},
  volume={7},
  number={ETRA},
  pages={1--18},
  year={2023},
  publisher={ACM New York, NY, USA}
}

@article{kubler2017subsmatch,
  title={SubsMatch 2.0: Scanpath comparison and classification based on subsequence frequencies},
  author={K{\"u}bler, Thomas C and Rothe, Colleen and Schiefer, Ulrich and Rosenstiel, Wolfgang and Kasneci, Enkelejda},
  journal={Behavior research methods},
  volume={49},
  number={3},
  pages={1048--1064},
  year={2017},
  publisher={Springer}
}

@inproceedings{abdrabou2025gaze,
  title={From gaze to data: Privacy and societal challenges of using eye-tracking data to inform GenAI models},
  author={Abdrabou, Yasmeen and {\"O}zdel, S{\"u}leyman and Maquiling, Virmarie and Bozkir, Efe and Kasneci, Enkelejda},
  booktitle={Proceedings of the 2025 Symposium on Eye Tracking Research and Applications},
  pages={1--9},
  year={2025}
}

@article{anderson2013recurrence,
  title={Recurrence quantification analysis of eye movements},
  author={Anderson, Nicola C and Bischof, Walter F and Laidlaw, Kaitlin EW and Risko, Evan F and Kingstone, Alan},
  journal={Behavior research methods},
  volume={45},
  number={3},
  pages={842--856},
  year={2013},
  publisher={Springer}
}

@inproceedings{reani2018investigation,
  title={An investigation of the effects of n-gram length in scanpath analysis for eye-tracking research},
  author={Reani, Manuele and Peek, Niels and Jay, Caroline},
  booktitle={Proceedings of the 2018 acm symposium on eye tracking research \& applications},
  pages={1--8},
  year={2018}
}

@article{yarbus1967eye,
  title={Eye Movements and Vision},
  author={Yarbus, Alfred L},
  journal={Eye movements and vision.},
  pages={171},
  year={1967},
  publisher={Springer US}
}

@inproceedings{zhang2015appearance,
  title={Appearance-based gaze estimation in the wild},
  author={Zhang, Xucong and Sugano, Yusuke and Fritz, Mario and Bulling, Andreas},
  booktitle={Proceedings of the IEEE conference on computer vision and pattern recognition},
  pages={4511--4520},
  year={2015}
}

@inproceedings{mondal2023gazeformer,
  title={Gazeformer: Scalable, effective and fast prediction of goal-directed human attention},
  author={Mondal, Sounak and Yang, Zhibo and Ahn, Seoyoung and Samaras, Dimitris and Zelinsky, Gregory and Hoai, Minh},
  booktitle={Proceedings of the IEEE/CVF conference on computer vision and pattern recognition},
  pages={1441--1450},
  year={2023}
}

@article{laborde2026vision,
  title={Vision toolkit part 3. Scanpaths and derived representations for gaze behavior characterization: a review},
  author={Laborde, Quentin and Roques, Axel and Armougum, Allan and Vayatis, Nicolas and Bargiotas, Ioannis and Oudre, Laurent},
  journal={Frontiers in Physiology},
  volume={16},
  pages={1721768},
  year={2026}
}

@inproceedings{kellnhofer2019gaze360,
  title={Gaze360: Physically unconstrained gaze estimation in the wild},
  author={Kellnhofer, Petr and Recasens, Adria and Stent, Simon and Matusik, Wojciech and Torralba, Antonio},
  booktitle={Proceedings of the IEEE/CVF international conference on computer vision},
  pages={6912--6921},
  year={2019}
}

@article{chuk2017having,
  title={Is having similar eye movement patterns during face learning and recognition beneficial for recognition performance? Evidence from hidden Markov modeling},
  author={Chuk, Tim and Chan, Antoni B and Hsiao, Janet H},
  journal={Vision research},
  volume={141},
  pages={204--216},
  year={2017},
  publisher={Elsevier}
}

@article{chuk2014understanding,
  title={Understanding eye movements in face recognition using hidden Markov models},
  author={Chuk, Tim and Chan, Antoni B and Hsiao, Janet H},
  journal={Journal of vision},
  volume={14},
  number={11},
  pages={8--8},
  year={2014},
  publisher={The Association for Research in Vision and Ophthalmology}
}

@article{chuk2020eye,
  title={Eye movement analysis with switching hidden Markov models},
  author={Chuk, Tim and Chan, Antoni B and Shimojo, Shinsuke and Hsiao, Janet H},
  journal={Behavior research methods},
  volume={52},
  number={3},
  pages={1026--1043},
  year={2020},
  publisher={Springer}
}

@inproceedings{vaidyanathan2014recurrence,
  title={Recurrence quantification analysis reveals eye-movement behavior differences between experts and novices},
  author={Vaidyanathan, Preethi and Pelz, Jeff and Alm, Cecilia and Shi, Pengcheng and Haake, Anne},
  booktitle={Proceedings of the symposium on eye tracking research and applications},
  pages={303--306},
  year={2014}
}

@article{anderson2015comparison,
  title={A comparison of scanpath comparison methods},
  author={Anderson, Nicola C and Anderson, Fraser and Kingstone, Alan and Bischof, Walter F},
  journal={Behavior research methods},
  volume={47},
  number={4},
  pages={1377--1392},
  year={2015},
  publisher={Springer}
}

@article{chen2021coco,
  title={Coco-search18 fixation dataset for predicting goal-directed attention control},
  author={Chen, Yupei and Yang, Zhibo and Ahn, Seoyoung and Samaras, Dimitris and Hoai, Minh and Zelinsky, Gregory},
  journal={Scientific reports},
  volume={11},
  number={1},
  pages={8776},
  year={2021},
  publisher={Nature Publishing Group UK London}
}

@inproceedings{hollenstein2020zuco,
  title={ZuCo 2.0: A dataset of physiological recordings during natural reading and annotation},
  author={Hollenstein, Nora and Troendle, Marius and Zhang, Ce and Langer, Nicolas},
  booktitle={Proceedings of the Twelfth Language Resources and Evaluation Conference},
  pages={138--146},
  year={2020}
}

@misc{maquiling2026motionbased,
      title={Motion-Based Tokenization for Cross-Dataset Egocentric Gaze Modeling}, 
      author={Virmarie Maquiling and Zhuojiang Cai and Enkelejda Kasneci},
      year={2026},
      eprint={2608.22926},
      archivePrefix={arXiv},
      primaryClass={cs.CV},
      url={https://arxiv.org/abs/2608.22926}, 
}

@article{foulsham2008turning,
  title={Turning the world around: Patterns in saccade direction vary with picture orientation},
  author={Foulsham, Tom and Kingstone, Alan and Underwood, Geoffrey},
  journal={Vision research},
  volume={48},
  number={17},
  pages={1777--1790},
  year={2008},
  publisher={Elsevier}
}

@article{nystrom2010adaptive,
  title={An adaptive algorithm for fixation, saccade, and glissade detection in eyetracking data},
  author={Nystr{\"o}m, Marcus and Holmqvist, Kenneth},
  journal={Behavior research methods},
  volume={42},
  number={1},
  pages={188--204},
  year={2010},
  publisher={Springer}
}

@inproceedings{kroger2020does,
  title={What does your gaze reveal about you? On the privacy implications of eye tracking},
  author={Kr{\"o}ger, Jacob Leon and Lutz, Otto Hans-Martin and M{\"u}ller, Florian},
  booktitle={IFIP International Summer School on Privacy and Identity Management},
  pages={226--241},
  year={2020},
  organization={Springer}
}

@article{deng2023eyettention,
  title={Eyettention: An attention-based dual-sequence model for predicting human scanpaths during reading},
  author={Deng, Shuwen and Reich, David R and Prasse, Paul and Haller, Patrick and Scheffer, Tobias and J{\"a}ger, Lena A},
  journal={Proceedings of the ACM on Human-Computer Interaction},
  volume={7},
  number={ETRA},
  pages={1--24},
  year={2023},
  publisher={ACM New York, NY, USA}
}

@inproceedings{sood2025scanez,
  title={ScanEZ: Integrating Cognitive Models with Self-Supervised Learning for Spatiotemporal Scanpath Prediction},
  author={Sood, Ekta and Dhar, Prajit and Troiano, Enrica and Southwell, Rosy and D’Mello, Sidney K},
  booktitle={Proceedings of the 63rd Annual Meeting of the Association for Computational Linguistics (Volume 2: Short Papers)},
  pages={1132--1142},
  year={2025}
}

@inproceedings{bolliger2023scandl,
  title={ScanDL: A diffusion model for generating synthetic scanpaths on texts},
  author={Bolliger, Lena and Reich, David and Haller, Patrick and Jakobi, Deborah and Prasse, Paul and J{\"a}ger, Lena},
  booktitle={Proceedings of the 2023 Conference on Empirical Methods in Natural Language Processing},
  pages={15513--15538},
  year={2023}
}

@inproceedings{fang2024oat,
  title={Oat: Object-level attention transformer for gaze scanpath prediction},
  author={Fang, Yini and Yu, Jingling and Zhang, Haozheng and van der Lans, Ralf and Shi, Bertram},
  booktitle={European Conference on Computer Vision},
  pages={366--382},
  year={2024},
  organization={Springer}
}

@article{marius2009gaze,
author = {Bernard Marius 't Hart and Johannes Vockeroth and Frank Schumann and Klaus Bartl and Erich Schneider and Peter König and Wolfgang Einhäuser},
title = {Gaze allocation in natural stimuli: Comparing free exploration to head-fixed viewing conditions},
journal = {Visual Cognition},
volume = {17},
number = {6-7},
pages = {1132--1158},
year = {2009},
publisher = {Routledge},
doi = {10.1080/13506280902812304},
URL = {https://doi.org/10.1080/13506280902812304},
eprint = {https://doi.org/10.1080/13506280902812304}
}

@article{land2006eye,
  title={Eye movements and the control of actions in everyday life},
  author={Land, Michael F},
  journal={Progress in retinal and eye research},
  volume={25},
  number={3},
  pages={296--324},
  year={2006},
  publisher={Elsevier}
}

@inproceedings{deng2023pre,
  title={Pre-trained language models augmented with synthetic scanpaths for natural language understanding},
  author={Deng, Shuwen and Prasse, Paul and Reich, David and Scheffer, Tobias and J{\"a}ger, Lena},
  booktitle={Proceedings of the 2023 Conference on Empirical Methods in Natural Language Processing},
  pages={6500--6507},
  year={2023}
}

@inproceedings{shubi2025decoding,
  title={Decoding reading goals from eye movements},
  author={Shubi, Omer and Hadar, Cfir Avraham and Berzak, Yevgeni},
  booktitle={Proceedings of the 63rd Annual Meeting of the Association for Computational Linguistics (Volume 1: Long Papers)},
  pages={5616--5637},
  year={2025}
}

@inproceedings{giulianelli2024proper,
  title={On the proper treatment of tokenization in psycholinguistics},
  author={Giulianelli, Mario and Malagutti, Luca and Gastaldi, Juan Luis and DuSell, Brian and Vieira, Tim and Cotterell, Ryan},
  booktitle={Proceedings of the 2024 Conference on Empirical Methods in Natural Language Processing},
  pages={18556--18572},
  year={2024}
}

@inproceedings{jarodzka2010vector,
  title={A vector-based, multidimensional scanpath similarity measure},
  author={Jarodzka, Halszka and Holmqvist, Kenneth and Nystr{\"o}m, Marcus},
  booktitle={Proceedings of the 2010 symposium on eye-tracking research \& applications},
  pages={211--218},
  year={2010}
}

@inproceedings{genzel2002entropy,
  title={Entropy rate constancy in text},
  author={Genzel, Dmitriy and Charniak, Eugene},
  booktitle={Proceedings of the 40th annual meeting of the Association for Computational Linguistics},
  pages={199--206},
  year={2002}
}

@inproceedings{wang2023does,
  title={Does listener gaze in face-to-face interaction follow the entropy rate constancy principle: An empirical study},
  author={Wang, Yu and Buschmeier, Hendrik},
  booktitle={Findings of the Association for Computational Linguistics: EMNLP 2023},
  pages={15372--15379},
  year={2023}
}

@inproceedings{wang2024much,
  title={How much does nonverbal communication conform to entropy rate constancy?: A case study on listener gaze in interaction},
  author={Wang, Yu and Xu, Yang and Skantze, Gabriel and Buschmeier, Hendrik},
  booktitle={Findings of the Association for Computational Linguistics: ACL 2024},
  pages={3533--3545},
  year={2024}
}

@inproceedings{genzel2003variation,
  title={Variation of entropy and parse trees of sentences as a function of the sentence number},
  author={Genzel, Dmitriy and Charniak, Eugene},
  booktitle={Proceedings of the 2003 conference on empirical methods in natural language processing},
  pages={65--72},
  year={2003}
}

\appendix

\section*{Appendix}
\label{sec:appendix}

\section{Implementation Details}
\label{app:impl_details}

The primary target-regret, n-gram-order, and pooling analyses use deterministic hash-based $80/10/10$ train/validation/test partitions over split seeds $0$--$4$. Fixed-seed structural diagnostics, sequence-level stability analyses, top-$k$ motif sensitivity, and the smoothing, interpolation, and VQDelta codebook-policy controls use seed $0$ unless otherwise stated. For each source$\rightarrow$target evaluation, the source-trained and target-trained models use the same tokenizer and n-gram configuration and are evaluated on the same target test sequences. Dataset sizes remain unequal, with AEA-fix being the smallest corpus.

AngDelta is a deterministic fixation-to-fixation motion tokenizer shared across all datasets. In the unified benchmark it uses $24$ yaw bins and $18$ pitch bins, yaw and pitch ranges of $[-120^\circ,120^\circ]$, bucketed outside handling, soft binning with temperature $\tau=1.0$, and hysteresis margin $0.05$. %
VQDelta is a learned but frozen motion tokenizer. The shared-vocabulary condition uses one global codebook with $K=128$ two-dimensional centroids, fit with k-means (\texttt{random\_state}=0, \texttt{n\_init=auto}) on at most $250{,}000$ normalized fixation-to-fixation deltas pooled from the training splits of all four datasets. Because the target training split contributes to this codebook, it is not a strict target-unseen condition. The codebook is frozen after fitting and reused for every directed transfer. The codebook-policy sensitivity analyses fit $K=128$ source-only and leave-target-out codebooks with codebook seed 0 and split seed 0.

For serialization, AngDelta emits a reserved initial zero entry (ID 1), while VQDelta emits an initial \texttt{<PAD>} entry (ID 0). Subsequent entries encode fixation-to-fixation displacements. The primary n-gram analyses operate on these complete serialized sequences.

The primary experimental configuration uses a count-based order-3 n-gram model with additive smoothing $k=0.05$ and recursive interpolation/backoff parameter $\gamma=10.0$. Target regret is the difference between source-trained and target-trained cross-entropy on the same target sequences, measured in bits per token. The target-trained model is treated as a finite-sample reference. Source-relative degradation, $(\mathrm{OOD}-\mathrm{IND})/\mathrm{IND}$, is reported only as a secondary descriptive quantity. Figure~\ref{fig:regime_transfer}B reports order sensitivity as the relative increase in target perplexity after within-sequence shuffling. The corresponding shuffle analysis reports the absolute shuffled-minus-real increase in target cross-entropy, in bits per token.

No dataset-specific hyperparameter search or test-set tuning was used. The primary experimental configuration fixes order 3, $k=0.05$, $\gamma=10.0$, the pooled $K=128$ VQDelta codebook, and top-200 motifs. Sensitivity analyses evaluate predefined n-gram-order, smoothing, interpolation, pooling, and codebook-policy controls. They are not used to select a configuration on the target test sets. The validation split is created by the shared split procedure, but the count-based transfer results do not use validation-based model selection.

Experiments were run on a Windows 11 workstation with an NVIDIA RTX 4090 GPU available. Because the unified fixation-level benchmark uses count-based order-3 n-gram models and a single frozen $K=128$ VQDelta codebook rather than neural sequence training, the main evaluation is lightweight and effectively CPU-bound. On this workstation, the core benchmark completed in under one hour, excluding upstream preprocessing and robustness analyses.

\begin{table*}[ht]
\centering
\small
\begin{tabular}{lllrrrrrrr}
\toprule
Dataset & Regime & Setting & Seq. & Train & Val. & Test & Tokens & Med. len. & Mean len. \\
\midrule
AEA-fix & Headworn & Everyday activities & 145 & 119 & 13 & 13 & 3017 & 11.0 & 20.8 \\
RITW-fix & Headworn & In-the-wild reading & 654 & 514 & 57 & 83 & 70321 & 72.0 & 107.5 \\
COCO & Screen & Visual search & 24880 & 19886 & 2508 & 2486 & 93823 & 3.0 & 3.8 \\
ZuCo & Screen & Sentence reading & 12847 & 10301 & 1281 & 1265 & 255064 & 17.0 & 19.9 \\
\bottomrule
\end{tabular}
\caption{Dataset statistics for the unified fixation-level benchmark. Counts are reported over exported fixation-level token sequences. All datasets use a deterministic hash-based 80/10/10 train/validation/test split with seed 0. The sequence counts are shared across AngDelta and VQDelta exports.}
\label{tab:dataset_stats}
\end{table*}

\begin{table*}[ht]
\centering
\small
\begin{tabular}{lll}
\toprule
Component & Setting & Value \\
\midrule
Data split & Policy & hash-based split on sequence ID \\
Data split & Fractions & 80/10/10 train/val/test \\
Data split & Seeds & 0--4 for target regret; 0 for fixed-seed diagnostics \\
AngDelta & yaw bins / pitch bins & 24 / 18 \\
AngDelta & yaw range / pitch range & $[-120^\circ,120^\circ]$ / $[-120^\circ,120^\circ]$ \\
AngDelta & outside policy & bucket \\
AngDelta & soft binning & on, $\tau=1.0$ \\
AngDelta & hysteresis margin & 0.05 \\
VQDelta & codebook size & $K=128$ \\
VQDelta & codebook training split & pooled train split over all four datasets \\
VQDelta & codebook seed & 0 \\
VQDelta & k-means settings & \texttt{random\_state}=0, \texttt{n\_init=auto} \\
VQDelta & max pooled deltas for fitting & 250{,}000 \\
Transfer model & n-gram order & 3 \\
Transfer model & additive smoothing & $k=0.05$ \\
Transfer model & interpolation/backoff & $\gamma=10.0$ \\
Diagnostics & motif top-$K$ & 200 in the main benchmark \\
Transfer model & model selection & fixed settings; validation split not used \\
Sensitivity & n-gram orders & 1, 2, 3, 4 \\
Sensitivity & additive smoothing $k$ & 0.01, 0.05, 0.10 at seed 0 \\
Sensitivity & interpolation $\gamma$ & 1, 10, 100 at seed 0 \\
VQDelta control & codebook policies & pooled, source-only, leave-target-out at $K=128$ \\
Stability & sequence bootstrap & 1000 replicates \\
\bottomrule
\end{tabular}
\caption{Implementation settings for the unified fixation benchmark.}
\label{tab:emnlp_impl_details}
\end{table*}

\section{Uncertainty over split seeds}
\label{app:uncertainty}

We repeated the target-regret evaluation over five deterministic hash-split seeds ($0$--$4$), while keeping the pooled $K=128$ VQDelta codebook fixed to isolate split variation from codebook variation. Mean order-3 regret was $2.150$ bits/token for AngDelta and $2.420$ for VQDelta. Regret was positive in 57/60 AngDelta and 58/60 VQDelta direction--seed evaluations. The weakest direction was RITW$\rightarrow$AEA, for which sequence-level bootstrap intervals included zero: $0.038$ bits/token with 95\% CI $[-0.307,0.465]$ for AngDelta and $0.133$ with 95\% CI $[-0.118,0.401]$ for VQDelta.

The descriptive transfer groupings do not show a general within-group advantage. Within-headworn regret is relatively low, but this average is partly driven by the uncertain RITW$\rightarrow$AEA direction, while within-screen regret is comparable to or larger than cross-group regret. Table~\ref{tab:unified_fixation_grouped_uncertainty} reports target regret and order sensitivity over split seeds. Motif summaries in Table~\ref{tab:unified_fixation_motif_uncertainty} quantify split-seed variation only. Motif sets were not bootstrapped by resampling complete sequences and are interpreted descriptively. Table~\ref{tab:absolute_transfer_ppl} presents absolute perplexity and source-relative degradation as secondary descriptive quantities.

\begin{table*}[ht]
\centering
\small
\begin{tabular}{llrr}
\toprule
Transfer regime & Tokenizer & Target regret & Order sens. \\
\midrule
headworn$\rightarrow$screen & AngDelta & $2.836 \pm 0.022$ & $1.250 \pm 0.027$ \\
screen$\rightarrow$headworn & AngDelta & $2.070 \pm 0.092$ & $0.857 \pm 0.158$ \\
within-headworn & AngDelta & $0.421 \pm 0.115$ & $1.087 \pm 0.127$ \\
within-screen & AngDelta & $2.667 \pm 0.018$ & $4.365 \pm 0.155$ \\
headworn$\rightarrow$screen & VQDelta & $3.071 \pm 0.006$ & $1.167 \pm 0.026$ \\
screen$\rightarrow$headworn & VQDelta & $2.455 \pm 0.140$ & $0.627 \pm 0.144$ \\
within-headworn & VQDelta & $0.453 \pm 0.099$ & $1.111 \pm 0.177$ \\
within-screen & VQDelta & $3.015 \pm 0.022$ & $4.441 \pm 0.125$ \\
\bottomrule
\end{tabular}
\caption{Target regret and order sensitivity by descriptive transfer regime over five split seeds. Values are means $\pm$ standard deviations over seed-level regime means. The regimes are descriptive because task, stimulus, device, preprocessing, representation, sequence length, and dataset size co-vary.}
\label{tab:unified_fixation_grouped_uncertainty}
\end{table*}

\begin{table*}[t]
\centering
\small
\begin{tabular}{llrr}
\toprule
Direction & Tok. & Bigram Jaccard & Trigram Jaccard \\
\midrule
aea\_fixation\_only->coco\_search18\_tp & ang\_delta & 0.064 $\pm$ 0.010 & 0.011 $\pm$ 0.004 \\
aea\_fixation\_only->ritw\_fixation\_only & ang\_delta & 0.501 $\pm$ 0.032 & 0.381 $\pm$ 0.021 \\
aea\_fixation\_only->zuco2 & ang\_delta & 0.192 $\pm$ 0.007 & 0.070 $\pm$ 0.006 \\
coco\_search18\_tp->aea\_fixation\_only & ang\_delta & 0.074 $\pm$ 0.009 & 0.017 $\pm$ 0.003 \\
coco\_search18\_tp->ritw\_fixation\_only & ang\_delta & 0.064 $\pm$ 0.006 & 0.022 $\pm$ 0.002 \\
coco\_search18\_tp->zuco2 & ang\_delta & 0.050 $\pm$ 0.006 & 0.015 $\pm$ 0.000 \\
ritw\_fixation\_only->aea\_fixation\_only & ang\_delta & 0.345 $\pm$ 0.058 & 0.179 $\pm$ 0.064 \\
ritw\_fixation\_only->coco\_search18\_tp & ang\_delta & 0.058 $\pm$ 0.008 & 0.011 $\pm$ 0.005 \\
ritw\_fixation\_only->zuco2 & ang\_delta & 0.234 $\pm$ 0.003 & 0.102 $\pm$ 0.005 \\
zuco2->aea\_fixation\_only & ang\_delta & 0.126 $\pm$ 0.030 & 0.048 $\pm$ 0.012 \\
zuco2->coco\_search18\_tp & ang\_delta & 0.048 $\pm$ 0.006 & 0.004 $\pm$ 0.002 \\
zuco2->ritw\_fixation\_only & ang\_delta & 0.213 $\pm$ 0.015 & 0.100 $\pm$ 0.010 \\
aea\_fixation\_only->coco\_search18\_tp & vq\_delta & 0.021 $\pm$ 0.008 & 0.001 $\pm$ 0.001 \\
aea\_fixation\_only->ritw\_fixation\_only & vq\_delta & 0.332 $\pm$ 0.049 & 0.126 $\pm$ 0.033 \\
aea\_fixation\_only->zuco2 & vq\_delta & 0.071 $\pm$ 0.008 & 0.001 $\pm$ 0.001 \\
coco\_search18\_tp->aea\_fixation\_only & vq\_delta & 0.018 $\pm$ 0.006 & 0.010 $\pm$ 0.004 \\
coco\_search18\_tp->ritw\_fixation\_only & vq\_delta & 0.021 $\pm$ 0.004 & 0.018 $\pm$ 0.003 \\
coco\_search18\_tp->zuco2 & vq\_delta & 0.005 $\pm$ 0.000 & 0.000 $\pm$ 0.000 \\
ritw\_fixation\_only->aea\_fixation\_only & vq\_delta & 0.185 $\pm$ 0.051 & 0.052 $\pm$ 0.026 \\
ritw\_fixation\_only->coco\_search18\_tp & vq\_delta & 0.017 $\pm$ 0.005 & 0.002 $\pm$ 0.001 \\
ritw\_fixation\_only->zuco2 & vq\_delta & 0.161 $\pm$ 0.005 & 0.001 $\pm$ 0.001 \\
zuco2->aea\_fixation\_only & vq\_delta & 0.048 $\pm$ 0.013 & 0.001 $\pm$ 0.001 \\
zuco2->coco\_search18\_tp & vq\_delta & 0.004 $\pm$ 0.001 & 0.000 $\pm$ 0.000 \\
zuco2->ritw\_fixation\_only & vq\_delta & 0.173 $\pm$ 0.014 & 0.000 $\pm$ 0.000 \\
\bottomrule
\end{tabular}
\caption{Motif-overlap uncertainty over split seeds for the unified fixation benchmark. Values are mean $\pm$ standard deviation across five split seeds.}
\label{tab:unified_fixation_motif_uncertainty}
\end{table*}

\begin{table*}[ht]
\centering
\small
\begin{tabular}{lllrrrr}
\toprule
Tokenizer & Source & Target & IND PPL & OOD PPL & Src.-relative degradation & Order sens. \\
\midrule
AngDelta & AEA-fix & COCO & 19.305 & 99.377 & 4.464 & 1.230 \\
AngDelta & AEA-fix & RITW-fix & 18.457 & 11.413 & -0.339 & 0.330 \\
AngDelta & AEA-fix & ZuCo & 19.197 & 55.102 & 2.039 & 0.504 \\
AngDelta & COCO & AEA-fix & 19.330 & 53.798 & 1.788 & 1.159 \\
AngDelta & COCO & RITW-fix & 19.330 & 34.887 & 0.805 & 0.251 \\
AngDelta & COCO & ZuCo & 19.330 & 86.588 & 3.480 & 1.178 \\
AngDelta & RITW-fix & AEA-fix & 7.094 & 19.792 & 1.787 & 1.844 \\
AngDelta & RITW-fix & COCO & 7.141 & 245.250 & 33.487 & 2.539 \\
AngDelta & RITW-fix & ZuCo & 7.136 & 85.020 & 10.941 & 0.725 \\
AngDelta & ZuCo & AEA-fix & 11.939 & 68.453 & 4.715 & 1.394 \\
AngDelta & ZuCo & COCO & 11.940 & 95.736 & 7.020 & 7.553 \\
AngDelta & ZuCo & RITW-fix & 11.939 & 46.744 & 2.916 & 0.625 \\
VQDelta & AEA-fix & COCO & 34.635 & 94.371 & 1.809 & 1.198 \\
VQDelta & AEA-fix & RITW-fix & 34.595 & 31.276 & -0.065 & 0.337 \\
VQDelta & AEA-fix & ZuCo & 34.629 & 119.026 & 2.537 & 0.420 \\
VQDelta & COCO & AEA-fix & 15.904 & 101.126 & 5.363 & 1.033 \\
VQDelta & COCO & RITW-fix & 15.904 & 113.091 & 6.118 & 0.144 \\
VQDelta & COCO & ZuCo & 15.904 & 194.357 & 11.224 & 1.078 \\
VQDelta & RITW-fix & AEA-fix & 17.115 & 36.088 & 1.119 & 1.885 \\
VQDelta & RITW-fix & COCO & 17.121 & 235.827 & 12.869 & 2.378 \\
VQDelta & RITW-fix & ZuCo & 17.120 & 153.276 & 7.998 & 0.672 \\
VQDelta & ZuCo & AEA-fix & 17.931 & 240.923 & 12.433 & 1.084 \\
VQDelta & ZuCo & COCO & 17.931 & 95.960 & 4.352 & 7.804 \\
VQDelta & ZuCo & RITW-fix & 17.931 & 120.001 & 5.687 & 0.247 \\
\bottomrule
\end{tabular}
\caption{Absolute in-domain and out-of-domain perplexities for the unified fixation benchmark. Values are multiseed means over split seeds 0--4. Source-relative degradation is presented as a secondary descriptive quantity and is not used for the primary transfer interpretation.}
\label{tab:absolute_transfer_ppl}
\end{table*}

\section{Robustness and Sensitivity Analyses}
\label{app:additional_analyses}
This appendix reports analyses of pooling strategy, codebook dependence, n-gram order, sequence shuffling, sequence length, and estimate stability.

\subsection{N-gram-order and shuffle controls}

Table~\ref{tab:order_controls} compares orders 1--4 over the five split seeds. Relative to order 1, order 3 reduces aggregate target cross-entropy by $0.392$ bits/token for AngDelta and $0.917$ for VQDelta. Order 2 has the lowest target cross-entropy for AngDelta, while orders 3 and 4 are nearly tied for VQDelta. Separately, shuffling tokens within each target sequence increases cross-entropy in all 24 directions for split seed 0, by a mean of $1.188$ bits/token for AngDelta and $1.113$ for VQDelta.

\begin{table}[ht]
\centering
\small
\setlength{\tabcolsep}{4pt}
\begin{tabular}{lrrrr}
\toprule
Tokenizer & Order & Target bits & Mean regret & Positive \\
\midrule
AngDelta & 1 & 6.776 & 1.941 & 60/60 \\
AngDelta & 2 & 6.232 & 2.030 & 58/60 \\
AngDelta & 3 & 6.384 & 2.150 & 57/60 \\
AngDelta & 4 & 6.460 & 2.111 & 57/60 \\
VQDelta & 1 & 7.934 & 2.674 & 60/60 \\
VQDelta & 2 & 7.216 & 2.521 & 58/60 \\
VQDelta & 3 & 7.017 & 2.420 & 58/60 \\
VQDelta & 4 & 7.013 & 2.374 & 58/60 \\
\bottomrule
\end{tabular}
\caption{N-gram-order controls over 12 directions and five split seeds. Target bits denotes source-trained cross-entropy on target sequences. Positive gives the number of direction--seed evaluations with positive target regret.}
\label{tab:order_controls}
\end{table}

At split seed 0, aggregate mean target regret remains positive for additive smoothing $k\in\{0.01,0.05,0.10\}$ and interpolation $\gamma\in\{1,10,100\}$. Exact likelihoods change across settings, so these controls support the sign of the aggregate transfer gap rather than invariance of its magnitude.

\subsection{Leave-target-out pooling}

For each target, the pooling controls combine the other three datasets. Table~\ref{tab:pooling_controls} reports the best of naive count pooling, dataset-balanced count pooling, and an equal-weight mixture of separately trained source-model probabilities. The mixture is best in six of eight conditions, balanced pooling in two, and naive pooling in none. Only AEA--AngDelta balanced pooling has negative mean regret.

\begin{table}[ht]
\centering
\small
\setlength{\tabcolsep}{4pt}
\begin{tabular}{lllr}
\toprule
Target & Tokenizer & Best multi-source strategy & Regret \\
\midrule
AEA & AngDelta & Balanced pooling & -0.113 \\
AEA & VQDelta & Probability mixture & 0.292 \\
COCO & AngDelta & Probability mixture & 2.102 \\
COCO & VQDelta & Probability mixture & 2.215 \\
RITW & AngDelta & Balanced pooling & 0.734 \\
RITW & VQDelta & Probability mixture & 1.138 \\
ZuCo & AngDelta & Probability mixture & 2.208 \\
ZuCo & VQDelta & Probability mixture & 2.400 \\
\bottomrule
\end{tabular}
\caption{Best leave-target-out multi-source strategy for each target and tokenizer over five split seeds. Regret is measured against the target-trained reference on the same target sequences.}
\label{tab:pooling_controls}
\end{table}

\subsection{VQDelta codebook policy}

Table~\ref{tab:vq_policy_controls} compares the pooled shared-vocabulary codebook with strict source-only and leave-target-out policies at $K=128$, codebook seed 0, and split seed 0. The policies produce comparable mean regret, but codebook refitting changes token assignments, occupancy, quantization, and transition statistics simultaneously. The differences therefore cannot be attributed only to the n-gram model.

\begin{table}[ht]
\centering
\small
\setlength{\tabcolsep}{3pt}
\begin{tabular}{lrrr}
\toprule
Codebook policy & Target bits & Mean regret & $\Delta$ target bits \\
\midrule
Pooled reference & 6.745 & 2.344 & 0.000 \\
Source-only & 6.575 & 2.232 & -0.170 \\
Leave-target-out & 6.465 & 2.283 & -0.280 \\
\bottomrule
\end{tabular}
\caption{VQDelta $K=128$ codebook-policy controls at split seed 0. The final column is relative to the pooled shared-vocabulary reference. Token identities are not aligned across separately fitted codebooks.}
\label{tab:vq_policy_controls}
\end{table}

\subsection{Sequence-level stability}

AEA contains 145 sequences with median length 11, but only six target-AEA test sequences remain after requiring length 12. COCO has median sequence length 3, only 1.25\% of sequences remain at length 12, and order-3 natural-context coverage is 37.3\%. Under natural-context-only scoring, RITW$\rightarrow$AEA regret becomes $-0.087$ bits/token for AngDelta and $-0.017$ for VQDelta. At minimum length 12, Spearman correlations of direction-level regret with the length-2 evaluation are $0.552$ and $0.559$, respectively. These controls support an uneven-transfer conclusion but not a stable pairwise ranking.

\section{Top-\texorpdfstring{$k$}{k} Motif Sensitivity}
\label{app:topk_sensitivity}

To test whether motif-overlap results depend on the choice of top-$k$, we recomputed bigram and trigram Jaccard overlap for $k \in \{25,50,100,200,500\}$ using the unified fixation-level benchmark. Table~\ref{tab:topk_motif_sensitivity} shows that the broad regime ordering is stable across thresholds: within-headworn overlap remains strongest, while cross-regime and within-screen overlaps remain lower.

\begin{table*}[ht]
\centering
\small
\setlength{\tabcolsep}{4pt}
\begin{tabular}{llrrrrr}
\toprule
Tokenizer & Regime family & $k=25$ & $k=50$ & $k=100$ & $k=200$ & $k=500$ \\
\midrule
\multicolumn{7}{l}{\textbf{Bigram Jaccard}} \\
\midrule
AngDelta & within-headworn    & 0.5090 & 0.4731 & 0.4667 & 0.4074 & 0.3061 \\
AngDelta & headworn$\rightarrow$screen & 0.0974 & 0.1247 & 0.1418 & 0.1361 & 0.1357 \\
AngDelta & screen$\rightarrow$headworn & 0.0909 & 0.1159 & 0.1203 & 0.1146 & 0.1165 \\
AngDelta & within-screen      & 0.0527 & 0.0365 & 0.0446 & 0.0444 & 0.0983 \\
\addlinespace
VQDelta & within-headworn    & 0.2825 & 0.2500 & 0.2499 & 0.2598 & 0.2796 \\
VQDelta & headworn$\rightarrow$screen & 0.0051 & 0.0102 & 0.0220 & 0.0643 & 0.0929 \\
VQDelta & screen$\rightarrow$headworn & 0.0000 & 0.0051 & 0.0115 & 0.0571 & 0.0760 \\
VQDelta & within-screen      & 0.0000 & 0.0000 & 0.0076 & 0.0178 & 0.0241 \\
\midrule
\multicolumn{7}{l}{\textbf{Trigram Jaccard}} \\
\midrule
AngDelta & within-headworn    & 0.2908 & 0.3554 & 0.2813 & 0.2439 & 0.2053 \\
AngDelta & headworn$\rightarrow$screen & 0.0211 & 0.0209 & 0.0391 & 0.0489 & 0.0497 \\
AngDelta & screen$\rightarrow$headworn & 0.0268 & 0.0211 & 0.0268 & 0.0406 & 0.0434 \\
AngDelta & within-screen      & 0.0000 & 0.0000 & 0.0000 & 0.0101 & 0.0111 \\
\addlinespace
VQDelta & within-headworn    & 0.1681 & 0.1003 & 0.0875 & 0.1036 & 0.0731 \\
VQDelta & headworn$\rightarrow$screen & 0.0000 & 0.0000 & 0.0000 & 0.0006 & 0.0033 \\
VQDelta & screen$\rightarrow$headworn & 0.0000 & 0.0000 & 0.0000 & 0.0000 & 0.0032 \\
VQDelta & within-screen      & 0.0000 & 0.0000 & 0.0000 & 0.0000 & 0.0000 \\
\bottomrule
\end{tabular}
\caption{Top-$k$ motif-overlap sensitivity in the unified fixation-level benchmark. Values are regime-averaged Jaccard overlaps for top-$k$ bigram and trigram motif sets. Across thresholds, within-headworn overlap remains the strongest regime for both tokenizers, while cross-regime and within-screen overlaps remain lower.}
\label{tab:topk_motif_sensitivity}
\end{table*}

\section{Fixation-only and Match-Length Controls}
\label{app:controls}

This appendix reports two construction-sensitivity analyses using source-relative degradation. The matched-length control truncates sequences to a common maximum length under the fixation-level interface, while the event-interface control compares fixation-only headworn exports with mixed-format fixation/saccade streams. These analyses show that degradation is sensitive to sequence length and event compatibility. Because degradation is also confounded by source and target predictability, the figures are treated as secondary construction-sensitivity diagnostics and are not used to establish a general regime ordering.

\begin{figure*}[ht]
    \centering
    \includegraphics[width=1\linewidth]{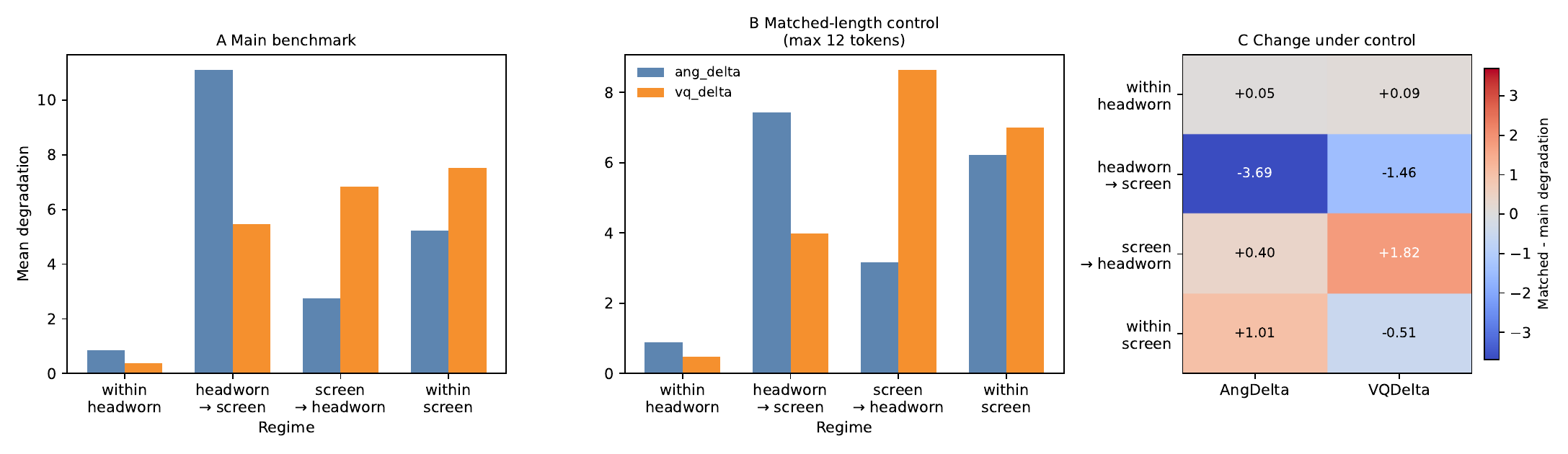}
    \caption{Matched-length source-relative-degradation control. All datasets use fixation-level exports, and sequences are capped at 12 tokens before evaluation. The changes in magnitude show that this secondary metric is sensitive to sequence construction. The figure is not used to infer a stable transfer-regime ordering.}
    \label{fig:matchedcontrol}
\end{figure*}

Figure~\ref{fig:matchedcontrol} summarizes the matched-length source-relative-degradation analysis. Truncating fixation-level sequences changes the reported values substantially in several directions, demonstrating that sequence construction affects this secondary metric. Target-regret sensitivity to minimum-length filtering is reported separately in Appendix~\ref{app:additional_analyses}.

\begin{figure*}[ht]
    \centering
    \includegraphics[width=.95\linewidth]{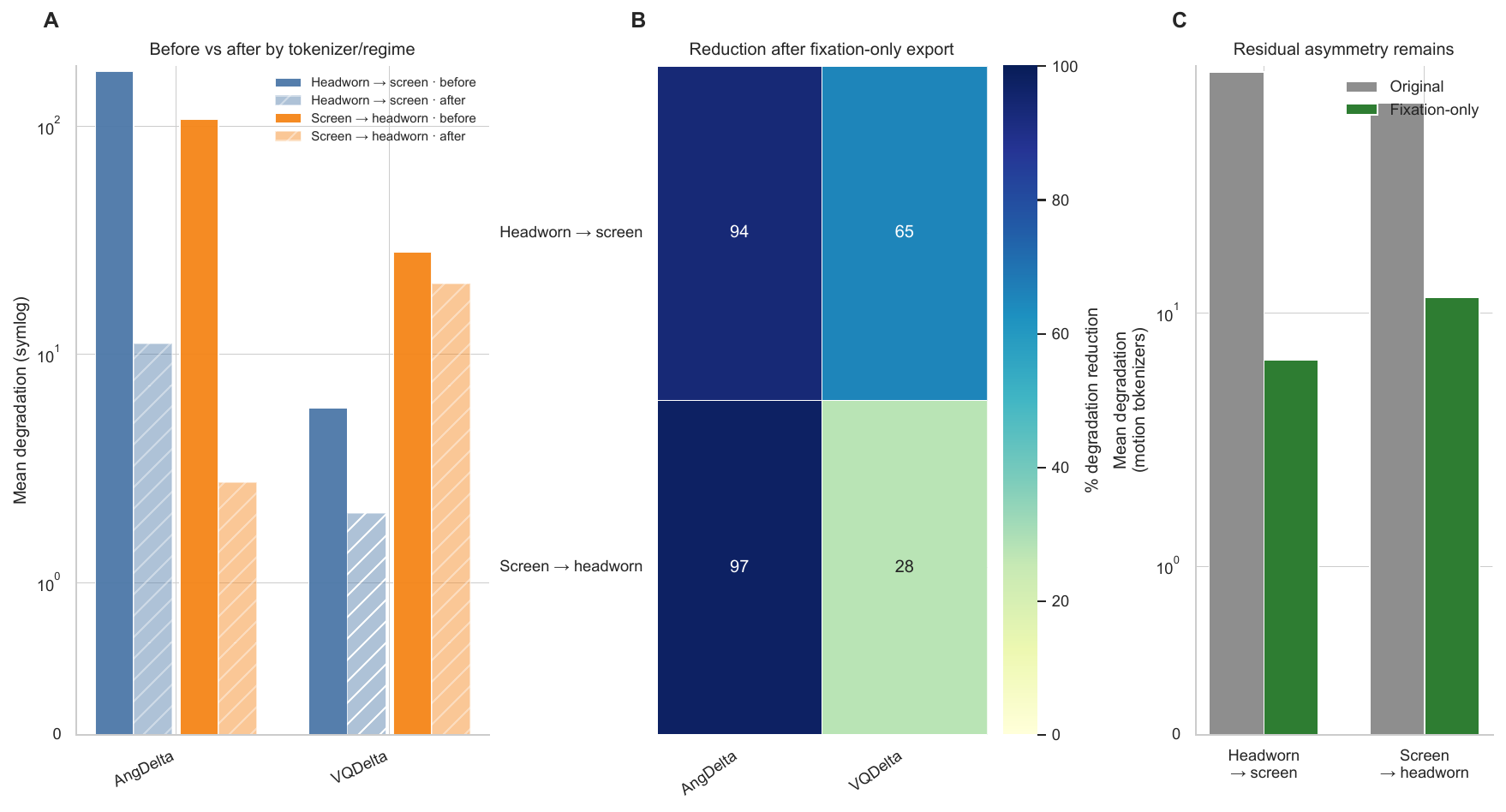}
    \caption{Effect of fixation-only conversion on source-relative degradation. The large changes demonstrate sensitivity to event-interface compatibility.}
    \label{fig:fixation_ablation}
\end{figure*}

Figure~\ref{fig:fixation_ablation} summarizes the event-interface control. Converting the headworn datasets to fixation-only sequences substantially changes cross-group degradation, confirming that incompatible event interfaces confound source-relative degradation. Residual directional differences remain, but they are descriptive and cannot be attributed causally to recording regime because task, stimulus, device, preprocessing, and dataset size also differ.

\section{Screen-Only Transfer: COCO and ZuCo}
\label{app:zucovscoco}

This appendix isolates the two screen-based datasets, COCO and ZuCo, to test whether sharing a screen-based fixation interface is sufficient for strong transfer. The comparison is useful because both datasets are fixation-native, but they differ strongly in task structure: goal-directed visual search versus sentence reading.

\begin{figure*}[ht]
    \centering
    \includegraphics[width=1\linewidth]{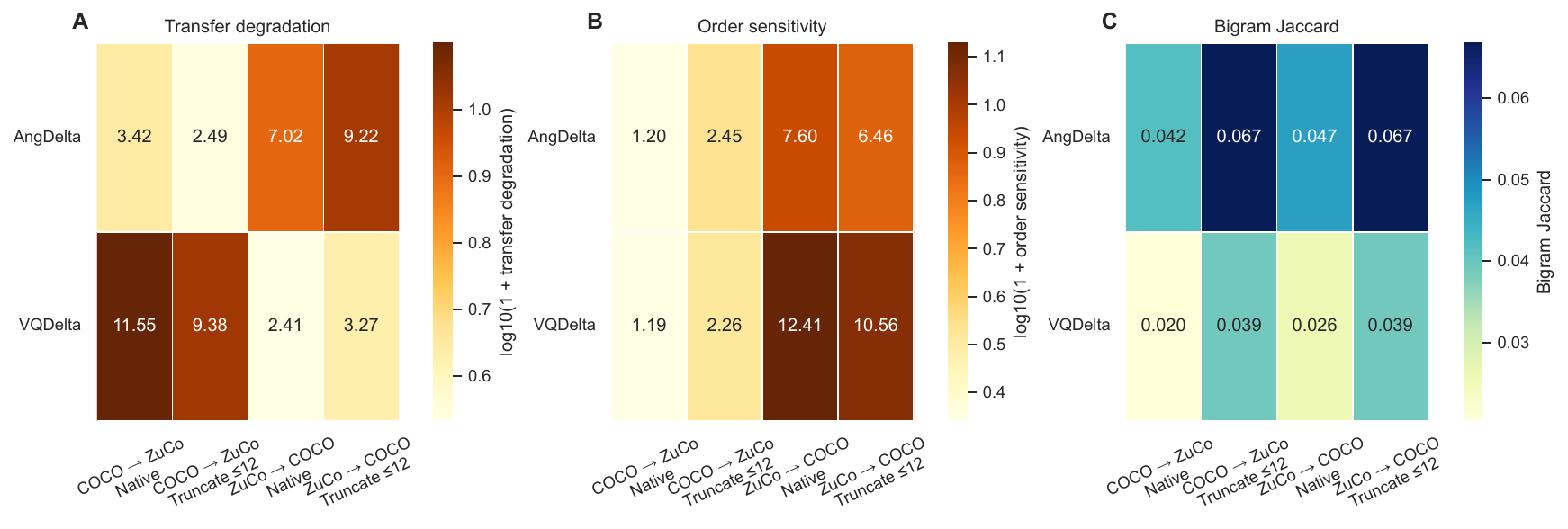}
    \caption{Descriptive COCO$\leftrightarrow$ZuCo comparison under native and truncate-to-12 evaluation. The panels report both directed transfers using secondary degradation and structural diagnostics.}
    \label{fig:screen_only_control}
\end{figure*}

Figure~\ref{fig:screen_only_control} provides a descriptive COCO$\leftrightarrow$ZuCo comparison. Truncation changes source-relative degradation and order-sensitivity values, while bigram overlap remains low under both sequence constructions. Because the datasets differ in task, stimulus, sequence length, and corpus size, this control does not isolate the cause of their differing transition statistics.

\end{document}